\documentclass[final,5p,times,twocolumn]{elsarticle}

\usepackage{hyperref}
\usepackage{xcolor}
\usepackage{amssymb}
\usepackage{subfigure}
\usepackage{graphicx}
\usepackage{acronym}

\newcommand{\DB}[1]{\emph{MultitaskPainting100k}}

\graphicspath{{./images/}}

\journal{Journal of \LaTeX\ Templates}

\begin{document}

\begin{frontmatter}

\title{Multitask Painting Categorization by Deep Multibranch Neural Network}

\author{Simone Bianco$^*$, Davide Mazzini, Paolo Napoletano, Raimondo Schettini}
\address{Department of Informatics, Systems and Communication (DISCo), University of Milano-Bicocca, Viale Sarca 336, 20126 Milan, Italy}

\cortext[mycorrespondingauthor]{Corresponding author}
\ead{\{bianco,napoletano,mazzini,schettini\}@disco.unimib.it}

\begin{abstract}

In this work we propose a new deep multibranch neural network to solve the tasks of artist, style, and genre categorization in a multitask formulation. In order to gather clues from low-level texture details and, at the same time, exploit the coarse layout of the painting, the branches of the proposed networks are fed with crops at different resolutions. We propose and compare two different crop strategies: the first one is a random-crop strategy that permits to manage the tradeoff between accuracy and speed; the second one is a smart extractor based on Spatial Transformer Networks trained to extract the most representative subregions. Furthermore, inspired by the results obtained in other domains, we experiment the joint use of hand-crafted features directly computed on the input images along with neural ones.

Experiments are performed on a new dataset originally sourced from wikiart.org and hosted by Kaggle, and made suitable for artist, style and genre multitask learning. The dataset here proposed, named \DB{}, is composed by 100K paintings, 1508 artists, 125 styles and 41 genres. Our best method, tested on the \DB{} dataset, achieves accuracy levels of 56.5\%, 57.2\%, and 63.6\% on the tasks of artist, style and genre prediction respectively.

\end{abstract}

\begin{keyword}
Painting Categorization, Painting Style Classification, Painter
Recognition, Deep Convolutional Neural Network, Multiresolution, Multitask
\MSC[2010] 00-01\sep  99-00
\end{keyword}

\end{frontmatter}

\section{Introduction}

\begin{figure*}[!ht]%
\centering
\resizebox{0.99\textwidth}{!}{
\includegraphics[width=1\textwidth]{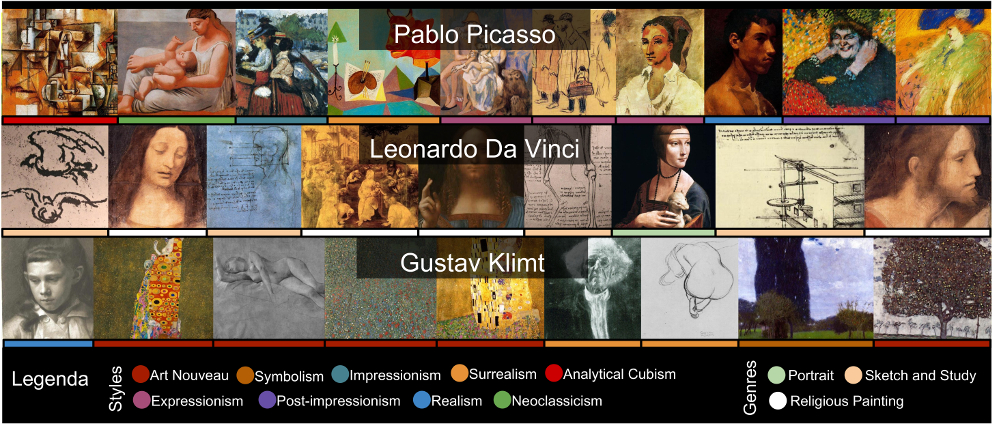}}
\caption{Paintings from the dataset adopted in this work, i.e. \DB{} dataset. Each row contains samples from a different artist. For each artist we show paintings with different genres and styles. Color coding is used to distinguish between genres and styles.}%
\label{fig:wikipaintingsivldataset}%
\end{figure*}

Automatic categorization and retrieval of digital paintings is gaining increasing attention due to the large quantities of visual artistic data made available by art museums that have digitized or are digitizing their artworks~\citep{carneiro2012artistic,mensink2014rijksmuseum,khan2014painting,mao2017deepart}. 

In this work, we deal with the problem of categorizing paintings by automatically predicting the artist who painted them (e.g. Monet, van Gogh, etc.), the pictorial styles (e.g. Impressionism, Baroque, etc.), and the genres (e.g. portrait, landscape, etc.)~\citep{anwer2016combining}. 
These three tasks are very challenging due to the large amount of both inter- and intra-class variations: in fact there are different personal styles in the same art movement, and the same artist may have drawn in one or more different pictorial styles and genres.
To have an idea of the difficulty of these tasks some examples taken from the dataset used in this work (i.e. \DB{}) are reported in Figure~\ref{fig:wikipaintingsivldataset}.

Artist classification consists in automatically associating the painting to its painter. In this task factors such as stroke patterns, the color palette used, the scene composition, and the subject depicted must be taken into account \citep{fichner2011foundations}.
Style classification consists in automatically assigning a painting into the school or art movement it belongs to. Art theorists define an artistic style as the combination of iconographic, technical and compositional features that give to a work its character \citep{widjaja2003identifying}. Style classification is complicated by the fact that styles may not remain pure but could be influenced by others.
Finally, genre classification consists in automatically categorizing a painting on the basis of the subject depicted.

\begin{table*}[!ht]

  \centering
  \caption{State of the art results for artist, style and genre categorization on the most used large scale painting datasets: Painting-91 \citep{khan2014painting}, WikiArt-WikiPaintings \citep{wikipaintings} and Art500k \citep{mao2017deepart}.}
  \label{tab:sota}
  \resizebox{1.3\columnwidth}{!}{
  \begin{tabular}{llllp{1.5cm}p{1.5cm}p{1.5cm}}
      \hline\noalign{\smallskip}

    &      & \multicolumn{2}{c}{Painting-91} & \multicolumn{3}{l}{WikiArt-WikiPaintings/Art500k}\\
    Model & Year & Artist & Style & Artist & Style & Genre \\
    \hline
    \cite{khan2014painting} &2014 & 53.1$_{(91)}$	& 62.2$_{(13)}$ \\
	\cite{peng2015cross} &2015 & 56.4$_{(91)}$	& 69.2$_{(13)}$	\\
    \cite{anwer2016combining} & 2016 & 64.5$_{(91)}$ &	78.4$_{(13)}$ \\
    \cite{chu2016deep} & 2016 & 63.2$_{(91)}$	& 73.6$_{(13)}$	& & 58.2$_{(25)}$ &\\
	\cite{puthenputhussery2016color} & 2016 & 59.0$_{(91)}$ & 67.4$_{(13)}$ \\
    \cite{puthenputhussery2016sparse} & 2016 & 65.8$_{(91)}$	& 73.2$_{(13)}$ \\
	\cite{peng2016toward} & 2016 & 57.3$_{(91)}$	& 70.1$_{(13)}$ \\
	\cite{banerji2016painting} & 2016 & 45.0$_{(91)}$ & 64.5$_{(13)}$ \\
    \cite{tan2016ceci} & 2016 & & & 76.1$_{(23)}$ & 54,5$_{(27)}$ & 74,1$_{(10)}$ \\
    \cite{salehlarge} & 2016 & & & 63.1$_{(23)}$ & 46.0$_{(27)}$ & 60.3$_{(10)}$\\
	\cite{bianco2017deeppainting} & 2017 & {78.5}$_{(91)}$ & {84.4}$_{(13)}$ \\
    \cite{huang2017fine} & 2017 & & & 81.9$_{(19)}$ & 50.1$_{(25)}$ & 69.0$_{(10)}$ \\
    \cite{mao2017deepart} & 2017 & & & 30.2$_{(1000)}$ & 39.2$_{(55)}$ & 39.2$_{(42)}$ \\
	\cite{chu2018image} & 2018 & 64.3$_{(91)}$ & 78.3$_{(13)}$ \\
  \hline
\end{tabular}
}
\end{table*}

The problems of automatic painter, style and genre categorization have been faced using different techniques. Some earlier approaches made use of traditional hand-crafted features \citep{carneiro2012artistic,khan2014painting} whereas more recent works rely on the use of deep neural networks for these tasks.

\cite{salehlarge} investigated a comprehensive list of visual features and metric learning approaches to learn an optimized similarity measure between paintings, which is then used to predict painting style, genre and artist.

\cite{anwer2016combining} used a deformable part model in order to combine low-level details and an holistic representation of the whole painting.
Inspired from the results obtained by deep networks as features extractors to solve different tasks \citep{sharif2014cnn,bianco2015local}, \cite{peng2015framework} used pretrained deep networks to deal with the small quantity of images available for painter and style categorization.
\cite{tan2016ceci} made different experiments by training a network from scratch or fine-tuning an existing network for the tasks of style and painter recognition.
Similarly, also \cite{banerji2016painting} investigated the use of a pre-trained network.
\cite{hentschel2016fine} performed interesting experiments about the quantity of data needed to fine-tune the network by \cite{krizhevsky2012imagenet} for the task of style classification.
\cite{chu2016deep,chu2018image} investigated the use of deep intra-layer and inter-layer correlation features as style descriptors, showing their superiority with respect to CNN features coming from fully-connected layers.
\cite{puthenputhussery2016color,puthenputhussery2016sparse} presented a novel set of image features that encode the local, color, spatial, relative intensity information
and gradient orientation of the painting image for painting artist classification, style classification as well as artist and style influence analysis.
\cite{peng2016toward} approached the problem of painter and style categorization together with other abstract tasks.

\cite{mao2017deepart} with the aim to generate a better representation of visual arts, presented a unified framework to learn joint representations that can simultaneously capture content and style of visual arts.
\cite{huang2017fine} proposed a novel two-channel deep residual network to classify fine-art painting images, where the first channel is the RGB channel and the second one is the brush stroke information channel.
\cite{bianco2017deeppainting} proposed a novel deep multibranch neural network to automatically predict painting’s artist and style, where the different branches processed the input image at different scales to jointly model the fine and coarse features of the painting.

All these works measure their performance mainly on three large scale datasets.
The most used is the Painting-91 dataset \citep{khan2014painting}, which consists of 4266 painting images from 91 different painters belonging to 13 different styles. This dataset is also the one used more consistently, since all the works adopting it use the same number of painters and styles.
Another large scale dataset is the WikiArt-WikiPaintings, that consists of 100,000 high-art images \citep{wikipaintings}. The dataset was built for the task of style recognition and originally from these images only the styles with more than 1,000 examples were selected, for a total of 25 styles and 85,000 images. Concerning the genre and artist recognition tasks, later works extracted from this dataset 10 different genres and from 19 to 23 artists.
The largest and most recent dataset is the Art500k \citep{mao2017deepart}, which contains 554,198 images of visual arts mainly scraped from WikiArt, Web Gallery of Art, Rijks Museum,
and Google Arts \& Culture websites. From these images 1,000 artists, 55 styles and 42 genres were extracted.

The average accuracy for the task of artist, style, and genre classification obtained by the state of the art approaches described, measured on the datasets respectively adopted are reported in Table \ref{tab:sota}. For each entry we also report as a subscript the number of classes considered in the experiments presented in each paper, that may be lower than the number of classes actually available in the original dataset.

This work builds on the results obtained in our previous work \citep{bianco2017deeppainting} and significantly extends it, adding the following main contributions: 
\begin{itemize}
\item[-] a novel deep neural network architecture, where different branches process the input image at different scales to jointly model fine and coarse features of the painting. The architecture is designed to simultaneously perform the classification of the author, the style and the genre in a multitask setup in order to both reduce the processing time and to induce a form of regularization;
\item[-] the use of a trainable crop strategy to feed the network branches with the most significant regions for painting categorization \citep{jaderberg2015spatial};
\item[-] the use of hand-crafted features for painting categorization; the best performing features have been exploited in our model by applying feature injection;
\item[-] a new dataset created starting from a dataset originally collected for a public competition on painter verification, and made suitable for artist, style and genre multitask learning. The final dataset is named \DB{}, and is composed of 100k paintings from 1508 artists, 125 styles and 41 genres;
\item[-] the evaluation of different strategies, with our best performing method achieving an accuracy level of 56.5\%, 57.2\%, and 63.6\% on the tasks of artist, style and genre prediction respectively on the \DB{} dataset considering all the 1508 artists, 125 styles and 41 genres.
\end{itemize}

\section{Deep Multibranch Neural Network}

\begin{figure*}[!htb]
\centering
\includegraphics[width=0.9\textwidth]{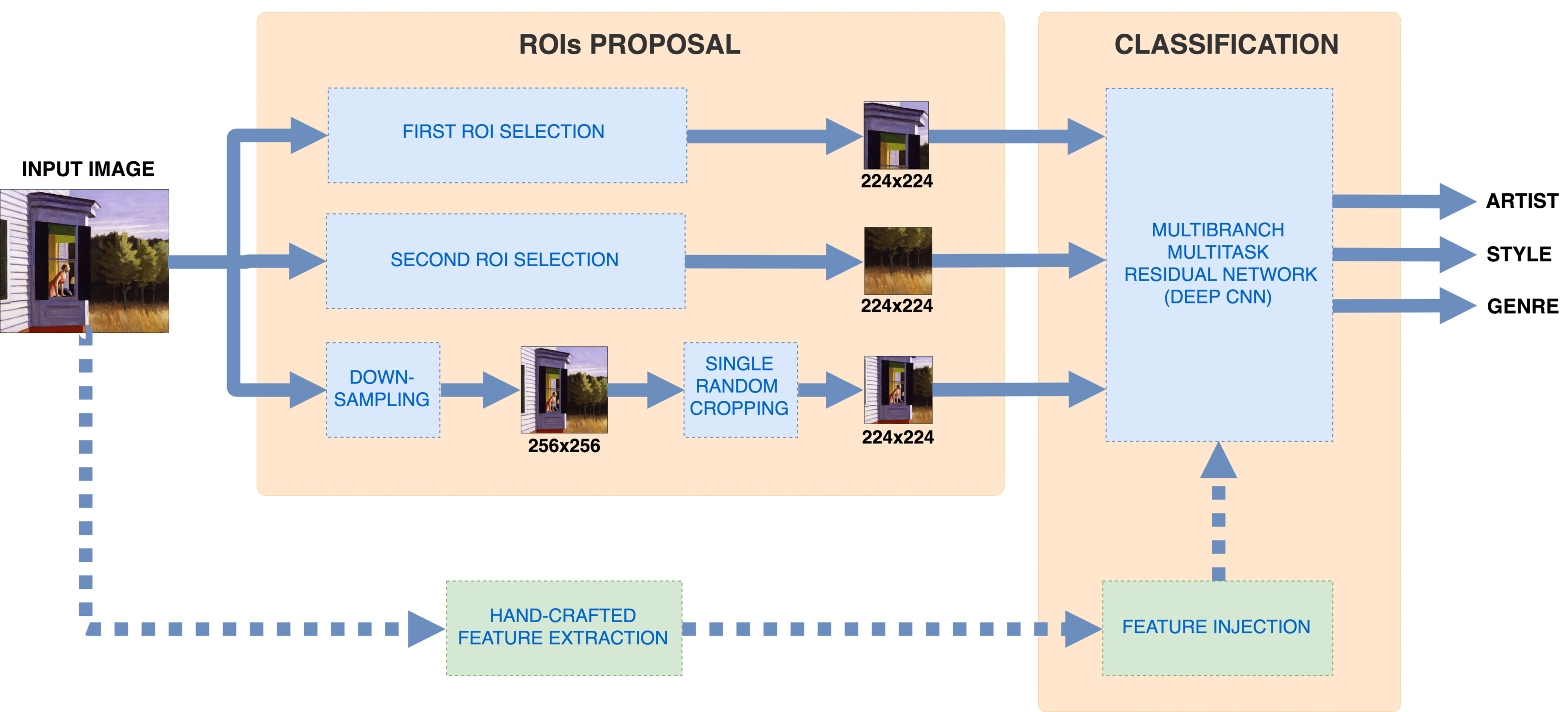}
\caption{Scheme of our Deep Multibranch Neural Network.}
\label{fig:network_structure}
\end{figure*}

Figure~\ref{fig:network_structure} shows the scheme of our Deep Multibranch Neural Network. The \emph{Regions Of Interest (ROIs) proposal} module  extracts three regions of interest from the input image that are sent into the multitask \emph{classification} module, which is made of a three-branch multitask residual network, and outputs the artist, style and genre prediction for the input image.

The scheme includes also the use of hand-crafted features, that are extracted directly from the input image and injected in the neural network before the last fully-connected layer. In the following sections each module of the scheme is discussed in more detail.

\subsection{ROIs proposal}
\label{sec:roisproposal}
The scene composition and the subject depicted are important clues to recognize a particular author or a painting style. These elements need to be extracted from the whole painting. At the same time finer details, such as stroke patterns or the line styles, are also very good clues.
Obviously a powerful discriminative model should consider both the coarse and fine level details. On the basis of these considerations we extract three subregions by following a multiresolution and multi-regions approach: two squared ``small'' subregions are extracted from the high-resolution image and one ``large'' subregion is extracted from the low-resolution image. We use only two scales since, in our preliminary experiments, the use of a higher number of scales brought a slight improvement compared to the exponential increase of computational burden.

Since paintings exhibit high variability in terms of aspect-ratios, the input images are resized such as the minimum side is 512 pixels and the aspect ratio is preserved. From the resulting images we extract two squared subregions of 224 by 224 pixels. Two possible ways to extract these two subregions are investigated: random  crop selection, and a trainable cropping strategy based on a Spatial Transformer Network.

The third subregion of 224 by 224 pixels is randomly selected from the images downsampled so that the minimum side is 256 pixels.

All the subregions extracted are squared, independently from the original aspect ratio of the input image. This is done to improve the computational efficiency of the GPU memory.
Images and regions sizes have been chosen as a trade-off between the resolution of fine details in smaller images and the computational burden of processing larger images.

\subsubsection{Random subregion selection}
The coordinates of the subregions inside the input image are randomly chosen with the only constraint that the selected subregions do not overlap. The rationale behind this choice is that the salient details can be anywhere inside the painting, and the extraction of subregions at no-overlapping random locations permits to increase the probability to get different painting details.

\subsubsection{Trainable subregion selection}
The subregions inside the input image are extracted by a trainable strategy that in the training phase learns how to extract the best subregions to maximize classification accuracy. The implemented strategy exploits a Spatial Transformer Network (STN), that was introduced by \cite{jaderberg2015spatial} to explicitly model the spatial manipulation of data within the network. The STN is composed by three modules. The first  module is a \emph{Localization Network}, that takes the input feature map $ U \in \mathbb{R}^{HxWxC}$ where $H,W,C$ represent the feature map width, height and channels respectively, and outputs the parameters $\theta$ of the transformation to be applied to the feature map, i.e. $\theta = f_{loc}(U)$.
The second module is a \emph{Parametrized Sampling Grid}, that takes as input the parameters from the Localization Network and produces a sampling grid.
The third module is the \emph{Bilinear Sampler}, which is a differentiable bilinear interpolation layer that takes as input the feature map and the sampling grid and performs the actual spatial warping.

In our implementation we used as Localization Network the ResNet-18 \citep{he2016deep} with the same type of Residual Blocks used for the main network as described in Section \ref{sec:netstructure}.
In order to maintain the geometric structure of the paintings, we also limited the type of transformations handled by the Sampling Grid layer allowing only translation and scale.

We used two STNs, one for each of the first two branches of our network, that are jointly trained with the rest of the network for multitask painting categorization.

\begin{table*}
  \centering
  \caption{Multibranch Multitask Deep Neural Network}
  \label{tab:npasses}
  \begin{tabular}{|c|c|c|c|}
  \hline
    Output Size & \multicolumn{3}{|c|}{Layers}\\
    \hline
  	 & branch 1 & branch 2 & branch 3\\
    \hline
    & Conv7 & Conv7 & Conv7\\
    & BatchNorm & BatchNorm & BatchNorm\\
        & ReLU & ReLU & ReLU\\
    112x112x64& MaxPool & MaxPool & MaxPool\\
    \hline
    56x56x256& 3$\times$ ResBlock& 3$\times$ ResBlock & 3$\times$ ResBlock\\
    \hline
    56x56x768& \multicolumn{3}{|c|}{Concatenation (channel dimension)} \\
    \hline
    56x56x256& \multicolumn{3}{|c|}{Join ResBlock, stride 1} \\
    \hline
    & \multicolumn{3}{|c|}{ResBlock, stride 2} \\
    28x28x512& \multicolumn{3}{|c|}{2$\times$ ResBlock} \\
        \hline
    & \multicolumn{3}{|c|}{ResBlock, stride 2} \\
    14x14x1024& \multicolumn{3}{|c|}{5$\times$ ResBlock} \\
    \hline
    & \multicolumn{3}{|c|}{ResBlock, stride 2} \\
    7x7x2048& \multicolumn{3}{|c|}{3$\times$ ResBlock} \\
    \hline
    1x1x2048& \multicolumn{3}{|c|}{AvgPool} \\
    \hline
    Num. Classes&FC-1508 & FC-125 & FC-41\\
    \hline
\end{tabular}
\end{table*}

\subsection{Classification: deep network architecture}\label{sec:netstructure}
A novel architecture based on Residual Blocks~\citep{he2016deep} that includes three branches and deals with the problem of artist, style and genre prediction at the same time is proposed. Table~\ref{tab:npasses} shows the architecture of our network more in detail. Each branch processes the subregions coming from the \emph{ROIs proposal} module separately until the processing flow is merged through the concatenation along the channel dimension of three $56 \times 56 \times 256$ feature maps to produce a $56 \times 56 \times 768$ feature map.

\begin{figure}
\centering
\includegraphics[width=0.4\columnwidth]{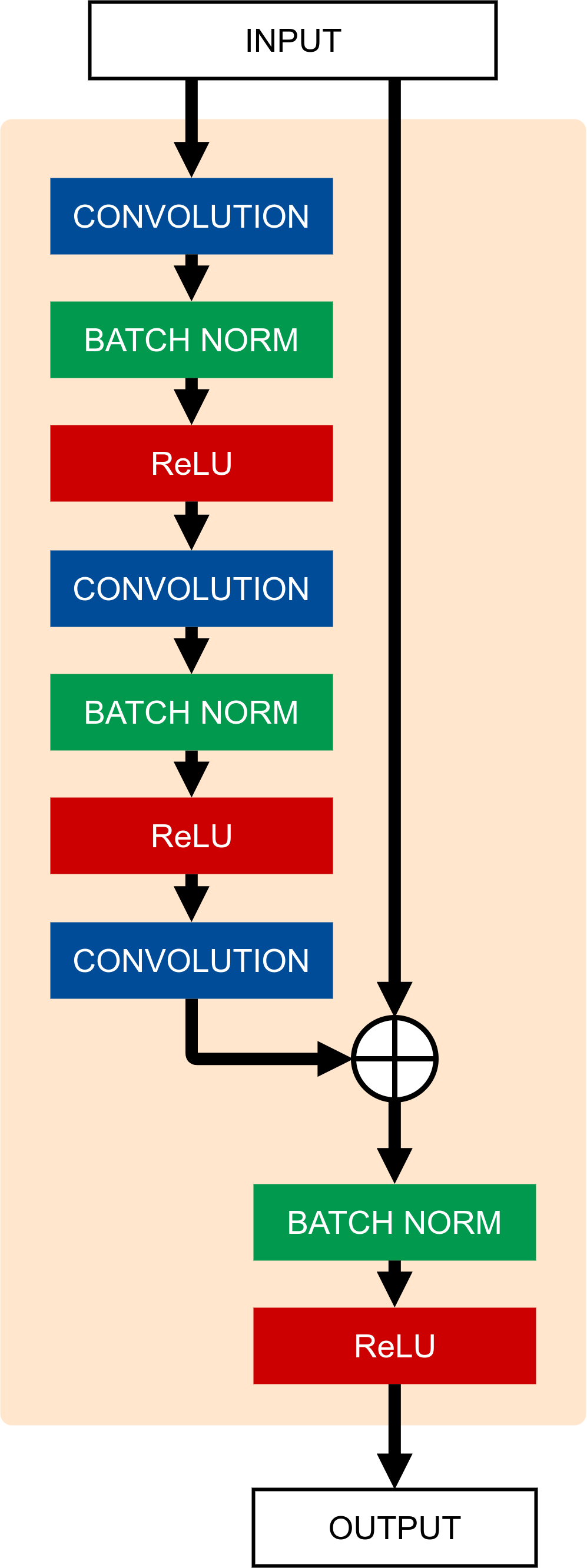}
\caption{The type of Residual Block used in our Deep Neural Network}
\label{fig:residual_block}
\end{figure}

Both in the three branches and in the classification module our deep network makes use of Residual Blocks which have been shown to be an effective architectural choice to build very deep networks \citep{he2016deep} and tackle the problem of vanishing gradients by using shortcut connections. In particular, we used ``bottleneck'' Residual Blocks,  which allow the network architecture to be even deeper \citep{he2016deep}. Each skip connection has four times the number of channels with respect to the internal elements of the block. This permits a large troughput of information among layers while maintaining a low computational complexity and low memory use inside each block. Our Residual Block structure is different from the one used by \cite{he2016deep}: we moved the Batch Normalization layer \citep{ioffe2015batch} after the sum with the skip connection because, in our experiments, the resulting configuration has shown better performances.

The Residual Block we used is shown in Figure \ref{fig:residual_block}. In our network

each of the three branches is composed by three Residual Blocks plus four layers near the input which perform the first processing (Convolution + BatchNorm \citep{ioffe2015batch} + ReLU \citep{nair2010rectified}) and an initial downsampling (Max Pooling).

The concatenation layer gathers the output of the three branches and stacks the output features along the channel dimension. The join layer is a particular Residual Block then converts the concatenated features to a smaller-dimensional feature space by compressing information along the channel dimension. The reason behind this operation is to make the computations feasible in the following layers by reducing the channel dimension of the output by a factor of three.

The common part of the network is composed by 13 Residual Blocks plus a spatial Average Pooling layer. While the Residual Blocks in the three branches do not include any downsampling operator, this part of the network includes convolution operators with stride two to perform a spatial downsampling of the input.

This leads to a gradual increasing of the receptive-field of the network in the deeper layers and also favors more abstract representations of the input.
In the final part of the classification module a fully-connected layer maps the output to the right number of classes depending on the task, respectively artist, style or genre prediction. Finally, the Fully-connected layer is followed by a Softmax layer that outputs the classes probabilities.

\subsection{Classification: hand-crafted feature injection}
We investigated the joint use of hand-crafted features along with learned neural features by adding them to the input of the last fully-connected layer of our network~\citep{bianco2017large}.

Hand-crafted descriptors are features extracted using manually predefined algorithms based on the expert knowledge. These features can be {global} and {local}~\citep{bianco2015local,napoletano2018visual}.

Global hand-crafted features describe an image as a whole in terms of colour, texture and shape distributions~\citep{mirmehdi2008handbook},  
while local hand-crafted descriptors, like Scale Invariant Feature Transform (SIFT)~\citep{lowe2004distinctive,bianco2015local}, provide a way to describe salient patches around properly chosen key points within the images. 

The hand-crafted features evaluated in this paper are the following:

\begin{itemize}
\item[-] 256-dimensional gray-scale histogram (Hist L)~\citep{novak1992anatomy};
\item[-] 768-dimensional RGB  histograms (Hist RGB)~\citep{Pietikainen1996};
\item[-] 10-dimensional feature vector composed of normalized
chromaticity moments, as defined in~\cite{paschos2003image} (Chromaticity);
  \item[-] 8-dimensional \emph{Dual Tree Complex Wavelet Transform}
 features obtained considering, for each color  channel, four scales, mean and
  standard deviation  (DT-CWT and DT-CWT L)~\citep{Bianconi2011,Barilla2008};
  \item[-] 512-dimensional \emph{Gist} features obtained considering eight
  orientations, four scales and 4 sub-windows  for each
  channel (Gist RGB)~\citep{oliva2001modeling};
\item[-] 32-dimensional \emph{Gabor} features composed of, for each color channel, mean and
  standard deviation of six orientations extracted at four frequencies, and normalized to be rotation invariant (Gabor L and Gabor RGB)~\citep{Bianconi2011,Bianconi2007};
\item[-] 243-dimensional \emph{Local Binary Patterns} (LBP) feature
  vector for each channel. We consider LBP applied to gray images
  and to color images represented in RGB~\citep{maenpaa2004classification}.  We select the LBP with a circular neighbourhood of radius 2 and 16 elements, and 18 uniform and no-rotation invariant patterns (LBP L and LBP RGB).
  \item[-] 499-dimensional LBP L combined with the Local Color Contrast (LCC) descriptor, as
  described in~\cite{cusano2016combining,cusano2014combining,cusano2013illuminant,bianco2013robustness}.
    \item[-] 144-dimensional {Colour and Edge Directivity Descriptor} (CEDD) features~\citep{chatzichristofis2008cedd}. This descriptor uses a fuzzy version of the five digital filters proposed by the MPEG-7 Edge Histogram Descriptor (EHD), forming 6 texture areas. CEDD uses 2 fuzzy systems that map the colours of the image in a 24-color custom palette;
 \item[-] 81-dimensional {Histogram of Oriented Gradients}
  feature vector~\citep{junior2009trainable}. Nine histograms with nine bins are concatenated to achieve the final feature vector  (HoG);

 \item[-] {1024}-dimensional \emph{Bag of Visual Words} (BoVW) of a 128-dimensional Scale Invariant Feature Transform (SIFT) calculated on the gray-scale image. The codebook of {1024} visual words is obtained by exploiting images from external sources~\citep{yang2010bag}.
\end{itemize}
The gray-scale image L is defined as follows:
L = 0.299R + 0.587G + 0.114B. All feature vectors have been $l^2$-normalized (i.e. they have been divided by their $l^2$-norm).

\section{Artist, style and genre: the \DB{} dataset}\label{sec:wikipaintingsivl}
The dataset used for the evaluation of our multitask deep multibranch neural network has been obtained from the Painter by Numbers Kaggle competition\footnote{https://www.kaggle.com/c/painter-by-numbers}. The goal of the competition was to predict if a pair of images are artworks made by the same artist or not. The dataset contained 103250 images of paintings  obtained mainly from WikiArt.org, that is a publicly available provider of digital artworks. Additional paintings have been provided by artists specifically for the competition. Images are at different resolutions but in general not smaller than 512px per side. The dataset includes a set of metadata for each painting, such as the artist name, style and genre of the painting. Giorgio De Chirico and Salvador Dal\'i are some examples of artist names. Romanticism and impressionism are some examples of painting styles, while cityscape and landscape are some examples of painting genres.

\begin{figure*}
\begin{tabular}{cc}
\multicolumn{2}{c}{\includegraphics[width=0.9\textwidth]{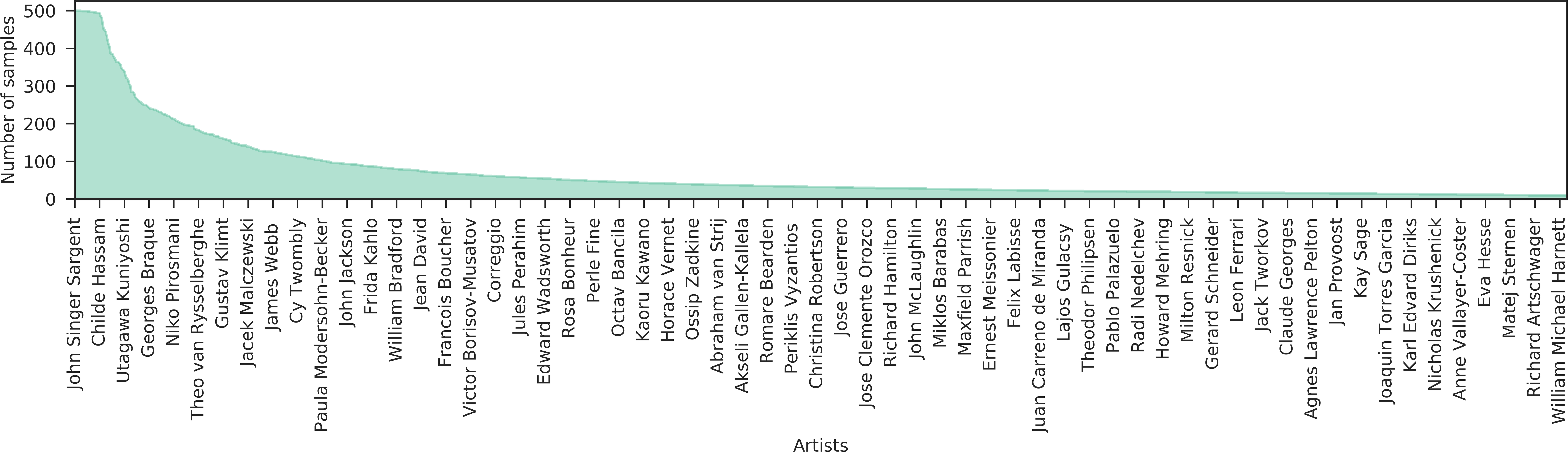}}\\
\multicolumn{2}{c}{(a)}\\[10pt]
\includegraphics[width=0.45\textwidth]{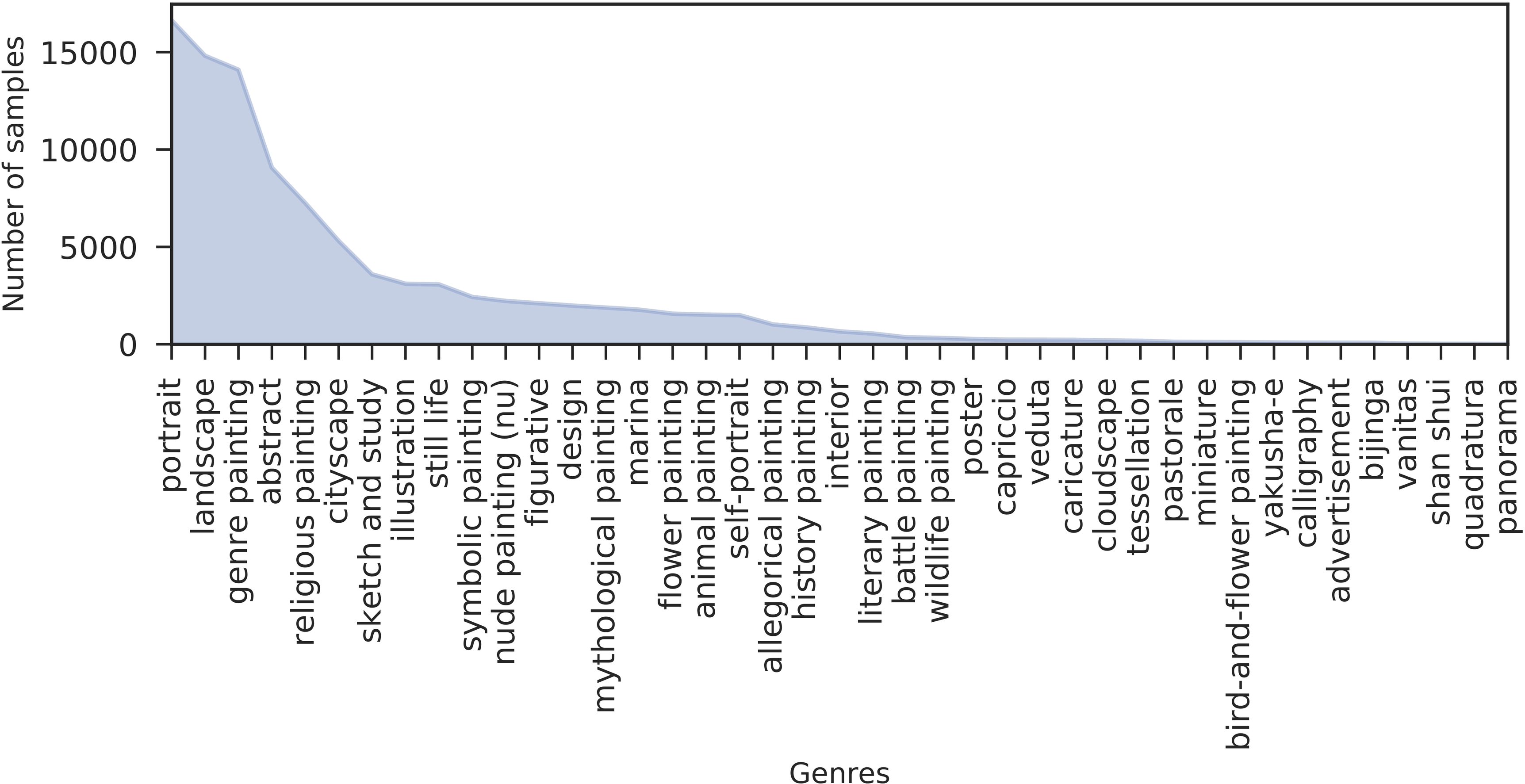}&\includegraphics[width=0.45\textwidth]{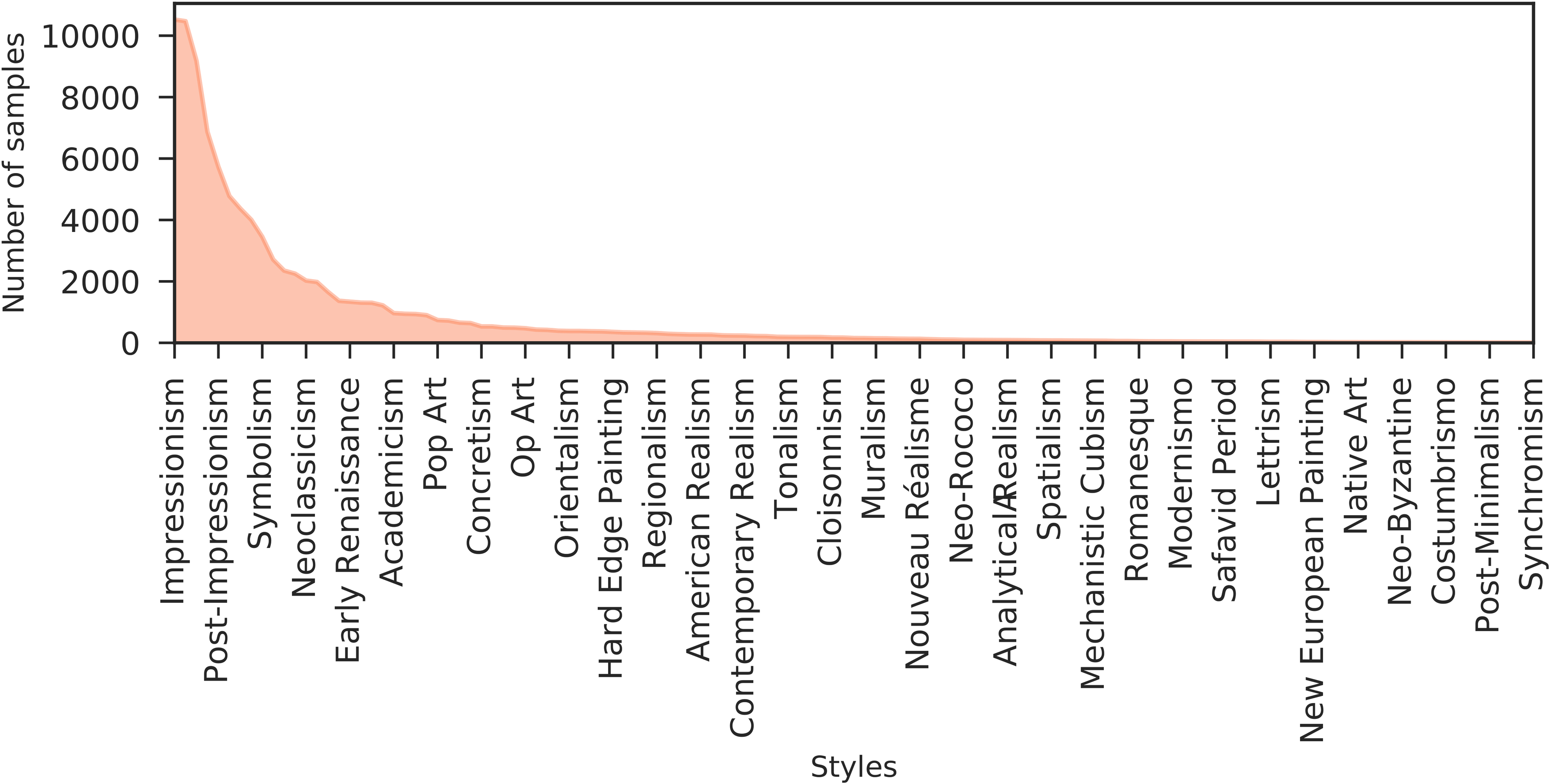}\\
(b)&(c)
\end{tabular}
\caption{Distributions of number of samples available for each of the 1508 artists (a), 41 genres (b) and 125 styles (c) within the \DB{} dataset. The names of classes are partially shown for lack of space. 
}
\label{fig:distributions}
\end{figure*}

While the competition provided a training/test split of the data to accomplish the task of predicting from a pair of images whether or not they are made by the same artist, we use this dataset here for another task: the prediction, given an image painting, of the artist name, style and genre. For this reason, the original split is not suitable for our task. To accomplish our task we select a subset of the original dataset such that there are at least 10 images in every class

for each of the three tasks, i.e. author, style and genre classification. After this selection the dataset contains 99816 images

for a total of
1508 artists, 125 styles and 41 genres. We call this selection the \DB{} dataset.
The dataset is split in two parts: a random 70\% belonging to the train set and the remaining 30\% to the test set.
We report in Fig.~\ref{fig:wikipaintingsivldataset} a subset of the paintings in the \DB{} dataset from three different artists (Pablo Picasso, Leonardo Da Vinci and Gustav Klimt) to let the reader getting the complexity of the recognition task. Each of the three selected artists has drawn, during his life, paintings with several styles and genres. This behavior is quite common among artists and this, along with the fact that the painting distribution is unbalanced across classes, makes recognition task quite challenging. Figures~\ref{fig:distributions}(a), (b) and (c) show the distributions of artists, genres and styles in terms of number of paintings for each class in the \DB{} dataset. In the case of the artist distribution it is clear that about 70\% of all artists have less than 100 paintings and about 50\% of the artists have less than 50 paintings. In the case of genres and styles we observe a similar behavior: 50\% of all the genres and styles have less than 1000 and 500 paintings respectively.

\begin{figure*}
\begin{tabular}{c}
\includegraphics[width=1\textwidth]{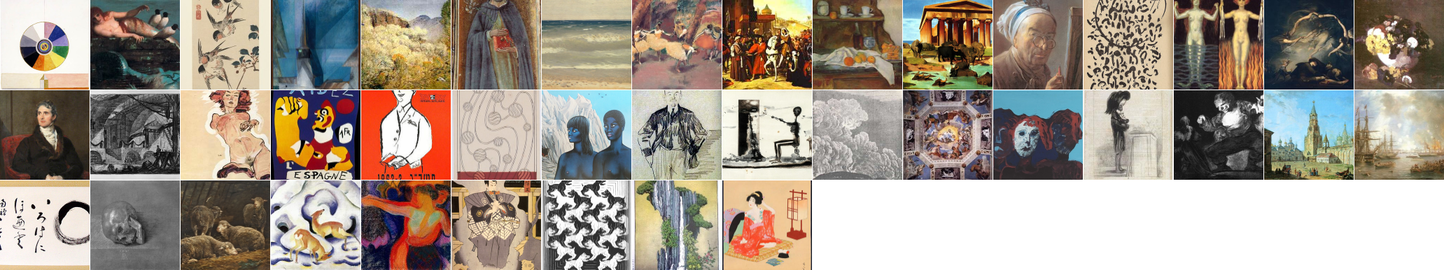}\\
(a)\\[10pt]
\includegraphics[width=1\textwidth]{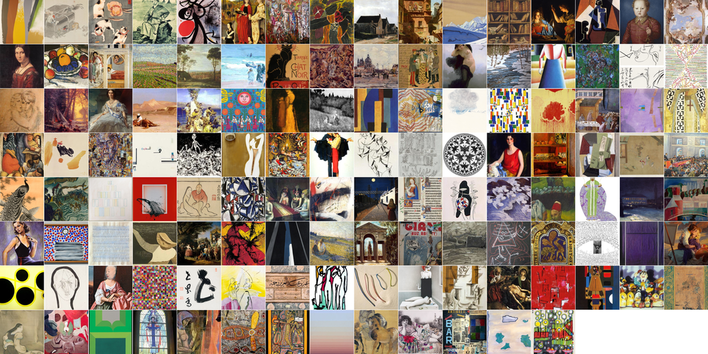}\\
(b)
\end{tabular}
\caption{Examples of each genre (a) and style (c) within the \DB{} dataset.}
\label{fig:genres_styles}
\end{figure*}

Figures~\ref{fig:genres_styles}(a) and (b) show a sample of each genre and style class within the \DB{} dataset. Images and annotations of the \DB{} dataset together with our train-test split will be made available on our website~\footnote{http://www.ivl.disco.unimib.it/activities/paintings/}.

\section{Experiments}
Our multitask deep multibrach neural network solutions is evaluated on the \DB{} dataset. We compare our solution with that from \cite{bianco2017deeppainting} that, as already shown in Table~\ref{tab:sota}, has demonstrated to perform much better than the state of the art on the Painting-91 dataset~\citep{khan2014painting}.

The solution in \cite{bianco2017deeppainting} included two different networks, one for the prediction of the artist name and another for the prediction of the painting style. To make it possible the comparison with the network proposed in this paper, we train three networks on the \DB{} dataset in order to accomplish the artist, style and genre prediction tasks separately.
In addition we compare also with the method by \citep{mao2017deepart}, that is the only work reporting results on a dataset having the closest number of classes to the proposed \DB{} dataset. Given the difficulty to exactly replicate their approach, we report the results taken from their paper.

In all the experiments, to cope with the small amount of training data we exploited some data augmentation techniques:
\begin{itemize}
  \item[-] Color jitter. It consists in randomly modifying contrast, brightness and saturation of the input image independently.
  \item[-] Lighting noise. It is a pixelwise transform based on the eigenvalues of the RGB pixel distribution of the dataset. It has been introduced by \cite{krizhevsky2012imagenet}.
  \item[-] Gaussian blur. It consists in applying a blur filter with fixed $\sigma$ to random images chosen with probability 0.5.
  \item[-] Geometric transforms. It includes small changes in scale and aspect-ratio of the input image.
\end{itemize}

\subsection{Results}
Table \ref{tab:wikipaintings_results} reports the comparison between the network proposed in~\cite{bianco2017deeppainting}, the one proposed in \cite{mao2017deepart}, and three variants of our proposal:
\begin{itemize}
\item[-] the multibranch-multitask network coupled with a random crop selection strategy;
\item[-] the multibranch-multitask network coupled with the trained crop selection strategy (STN);
\item[-] the multibranch-multitask network with the injection of HOG features, and coupled with the trained crop selection strategy (STN).
\end{itemize}
The performance is measured on the \DB{} dataset in terms of average classification accuracy, that is the mean of the accuracy obtained for each class.  

Looking at the results reported in Table~\ref{tab:wikipaintings_results} it is quite clear that the joint multitask training over all the tasks gives a big boost on style accuracy at the expense of a small decrease in performance on genre. The same happens with the injection of HOG features. We suppose that artist and style are much more correlated tasks, thus the training can benefit more from a joint loss optimization.
The use of Spatial Transformer Networks improves the performances on all tasks showing the contribution of the smart crop extraction strategy. Although a direct comparison with other state of the art method is not possible, we can indirectly compare our solution with that of  \cite{mao2017deepart}, that uses a dataset with a common origin to the \DB{} and has a similar number of classes. We can observe that for the artist task, even if we have 508 more classes, our results are 23.3\% better; for the style task, even if we have 70 more classes, our results are 18.0\% better; for the Genre task, where the results are more comparable since we have just one less class, our results are 24.4\% better.

\begin{table*}

  \centering
  \caption{Classification accuracy for the three different tasks on \DB{} dataset. Different models exploiting multitask, Spatial Transformer Networks (STN) and the injection of HOG features.}
  \label{tab:wikipaintings_results}
  \begin{tabular}{lccclcc}
    \hline\noalign{\smallskip}
     &  &  & \multicolumn{4}{c}{\DB{}} \\
    Model & Crop strategy & Feat. injection &Artist & Style & Genre & Average \\
    \hline
    \cite{mao2017deepart}$^\dag$ & & & 30.2$_{(1000)}$ & 39.2$_{(55)}$ & 39.2$_{(42)}$ & 36.2 \\
    Multibranch \citep{bianco2017deeppainting} & random & - & 53.1$_{(1508)}$ & 51.5$_{(125)}$ & \textbf{64.3}$_{(41)}$ & 56.3\\
    Multibranch multitask & random & - & 53.3$_{(1508)}$ & 55.4$_{(125)}$ & 63.0$_{(41)}$ & 57.2\\
    Multibranch multitask & STN & - & 56.1$_{(1508)}$ & 57.0$_{(125)}$ & 64.1$_{(41)}$ & \textbf{59.1}\\
    Multibranch multitask & STN & HOG & \textbf{56.5}$_{(1508)}$ & \textbf{57.2}$_{(125)}$ & 63.6$_{(41)}$ & \textbf{59.1}\\
  \hline
  $^\dag$ \footnotesize{Evaluated on the Art500k dataset}
\end{tabular}
\end{table*}

\subsection{Evaluation of hand-crafted features}
\begin{figure}
\centering
\includegraphics[width=1\columnwidth]{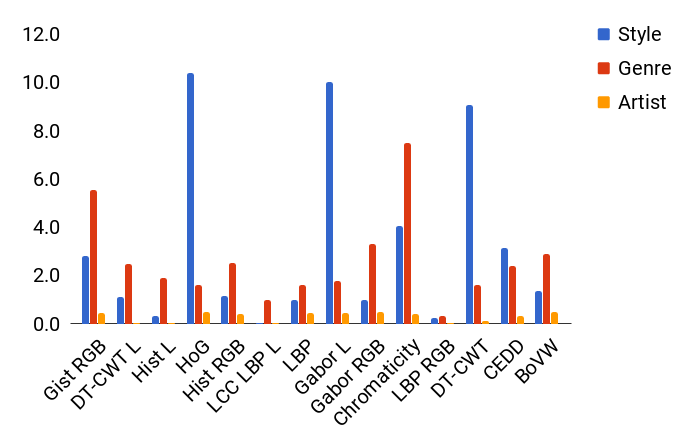}
\caption{Calssification accuracy (percentage). Hand-crafted features combined with a linear classifier to solve the three classification tasks.}
\label{fig:handcrafted_accuracy}
\end{figure}

To assess the improvement that hand-crafted features could bring to our existing architecture we did some preliminary experiments. We trained a linear classifier on top of each hand-crafted feature in order to classify each of the three tasks: artist, genre and style.
Figure \ref{fig:handcrafted_accuracy} shows the percentage of accuracy achieved by each hand-crafted feature and for each task.
This experiment gives a first glance on the discriminative power of the considered features for our classification tasks. As expected the accuracy for the task of artist prediction is quite low. This is the most difficult of the three tasks due to the large set of classes (i.e. 1508). On style and genre prediction some descriptors show an accuracy over 4\%. In particular the best features for style prediction are HOG and Gabor L, both grayscale descriptors, whereas for genre prediction genre classification the best descriptors are GIST color and chromaticity moments which relies both strongly on color information.

\begin{figure}
\centering
\includegraphics[width=1\columnwidth]{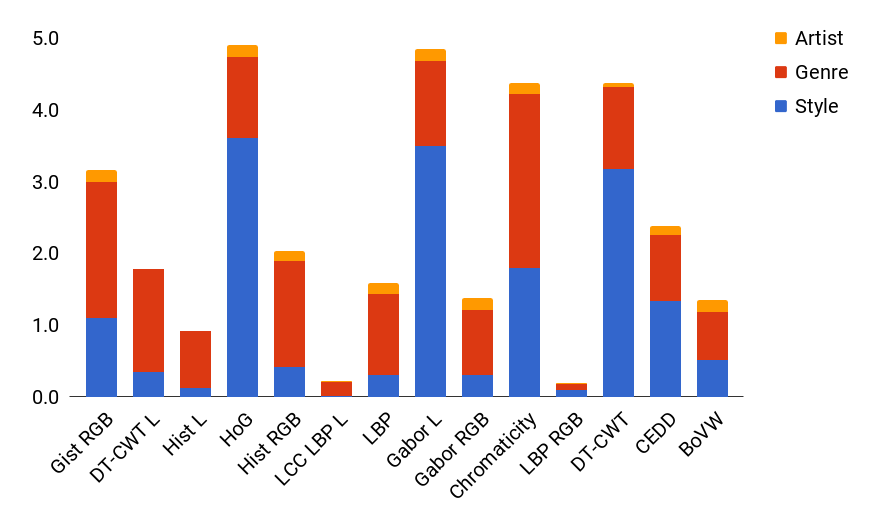}
\caption{Percentage of correctly classified examples by the hand-crafted features considered out of all misclassified examples by our multibranch multitask neural network with STN cropping strategy. Stacked bar chart for the three tasks together. HOG and Gabor L give the highest improvement.}
\label{fig:handcrafted_improvement}
\end{figure}

We made a second experiment in order to evaluate whenever an hand-crafted descriptor is able to correctly classify examples that are misclassified by our deep architecture. We used the trained linear classifiers on top of hand-crafted features to classify only the misclassified examples of our neural network architecture. 
Figure \ref{fig:handcrafted_improvement} shows a stacked bar graph. Each bar represents the cumulative contribution for all of the three tasks. From this graph are clearly visible the features that correctly classify the highest number of examples: HOG, Gabor L, Chromaticity Moments and DT-CWT.

These preliminary experiments help to highlight that among the hand-crafted descriptors, HOG is the most promising to be included in our classification pipeline. We fed the extracted features directly before the last fully-connected layer of our deep network. Table \ref{tab:wikipaintings_results} shows the accuracy achieved by the deep network combined with the HOG descriptor for the three classification tasks. In the case of artist and style classification, the use of HOG slightly improves the accuracy achieved by the deep network. In contrast, in the case of genre classification, the use of HOG features does not produce any improvements, so that the average accuracy over the three tasks is exactly the same with and without HOG features.


\subsection{Similarity search}

\begin{figure}
\centering
\includegraphics[width=1\columnwidth]{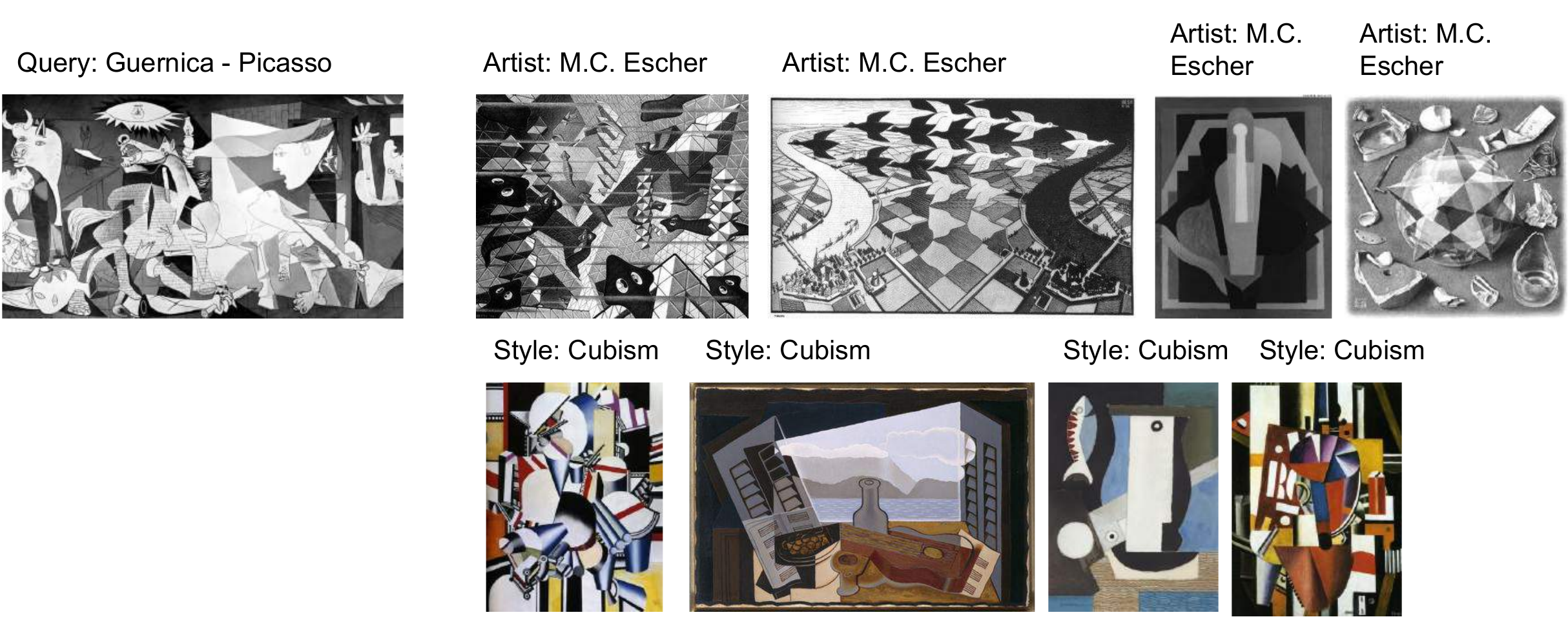}
\caption{Similarity results for {\it{Guernica}} by Picasso, belonging to the Cubism style. Similarity results: top four painting retrieved using the artist features (first row) and the style features (second row).}
\label{fig:example1}
\end{figure}

In the following we show some interesting results obtained when the proposed method is used for similarity search. Given one painting as input we extract three sets of features, one for each classification task faced (i.e. artist, style and genre). The features are the $l^2$-normalized activations of the last Fully-connected layer before the Softmax layer. In this way, each set of features can be used to compute the similarity in terms of artist, style and genre respectively.
We report the similarity results for four different paintings. For each of them, we retrieve the four most similar paintings using the artist features, and the four most similar paintings using the style features.
In Figure \ref{fig:example1} the {\it{Guernica}} painting by Picasso is fed to the system. The first four most similar paintings retrieved by the system using the artist features all belong to M.C. Escher, that in fact are much more similar to the input  than any other painting from Picasso himself. This example shows the difficulty of the task of painter recognition, especially for those artist that have used many different styles across their artistic production as for example Picasso himself (have a look to the first row of Figure \ref{fig:wikipaintingsivldataset} to see some examples). On the other hand, all the first four paintings retrieved using the style features belong to the Cubism style, that is the same style of the input painting.

\begin{figure}
\centering
\includegraphics[width=1\columnwidth]{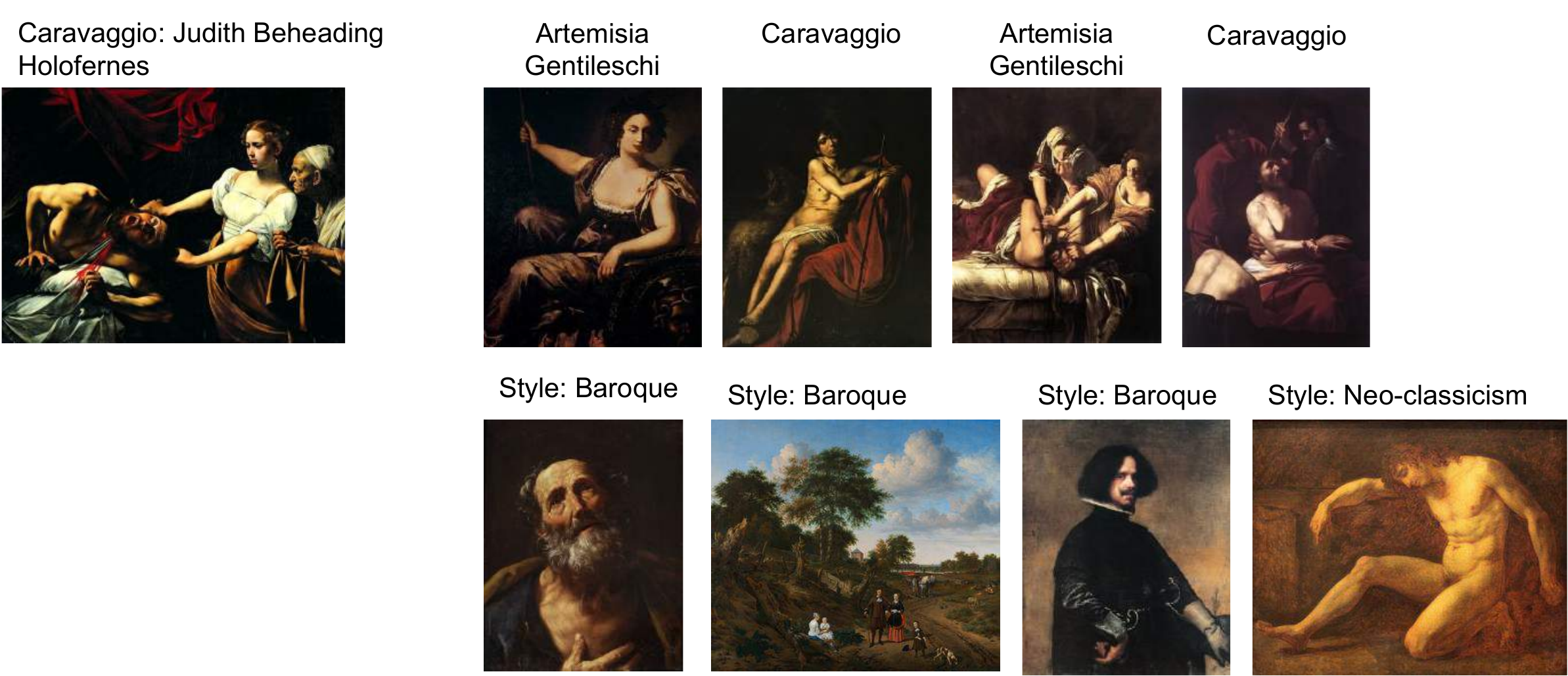}
\caption{Similarity results for {\it{Judith beheading Holofernes}} by Caravaggio, belonging to the Baroque style. Similarity results: top four painting retrieved using the artist features (first row) and the style features (second row).}
\label{fig:example2}
\end{figure}

A second example is reported in Figure \ref{fig:example2}, where the painting {\it{Judith beheading Holofernes}} by Caravaggio is given as input to our system. Although the most similar painting retrieved with the artist features is not from the correct author, the second and the fourth ones are from Caravaggio. Furthermore, it is interesting to notice that the third most similar painting, although from a different author, depicts the same subject, i.e. Judith beheading Holofernes, with very similar body poses. For what concerns the style, all the 
the first four paintings retrieved using the style features belong to the Baroque style, that is the same style of the input painting.

\begin{figure}
\centering
\includegraphics[width=1\columnwidth]{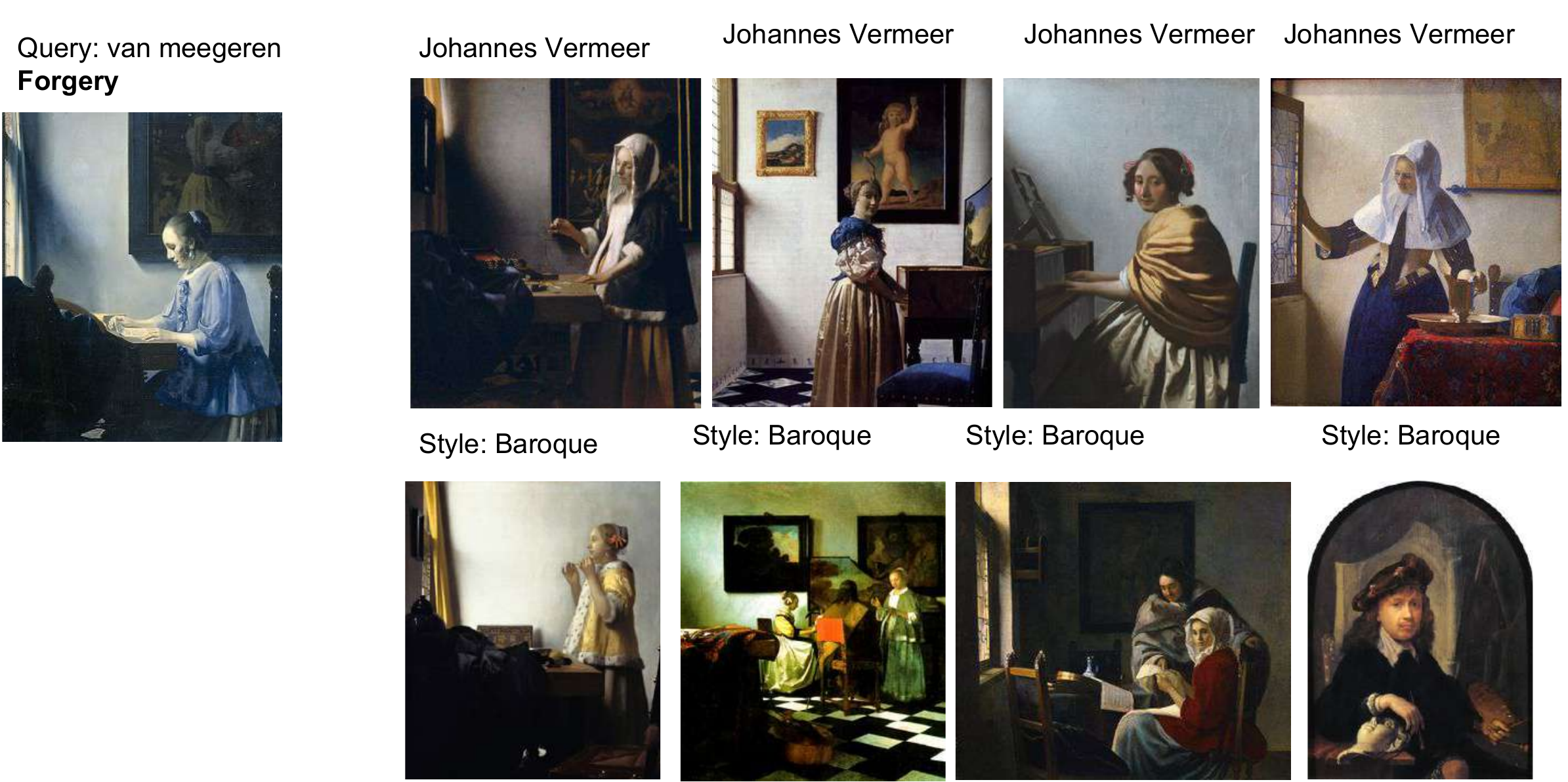}
\caption{Similarity results for a forged painting by van Meegeren, belonging to the Baroque style. Similarity results: top four painting retrieved using the artist features (first row) and the style features (second row).}
\label{fig:example3}
\end{figure}

A third example is reported in Figure \ref{fig:example3}, where a painting by H.A. van Meegeren (that does not belong to the \DB{} dataset), is given as input to the system. H.A. van Meegeren is famous for having forged paintings of some of the world's most famous artists, including Frans Hals, Pieter de Hooch, Gerard ter Borch, and Johannes Vermeer. He so well replicated the styles and colors of the artists that the best art critics and experts of the time regarded his paintings as genuine and sometimes exquisite. Then, it makes completely sense that the all the four most similar paintings retrieved by our system using the artist features are from Johannes Vermeer. Furthermore, the retrieved paintings are similar to the input also from a compositional point of view, with a girl painted against a clear wall, close to a table and with the same light coming from a window in the upper left corner of the painting. Concerning artistic style, 
all the the first four paintings retrieved using the style features belong to the Baroque style, that is the same style of the input painting.

\begin{figure}
\centering
\includegraphics[width=1\columnwidth]{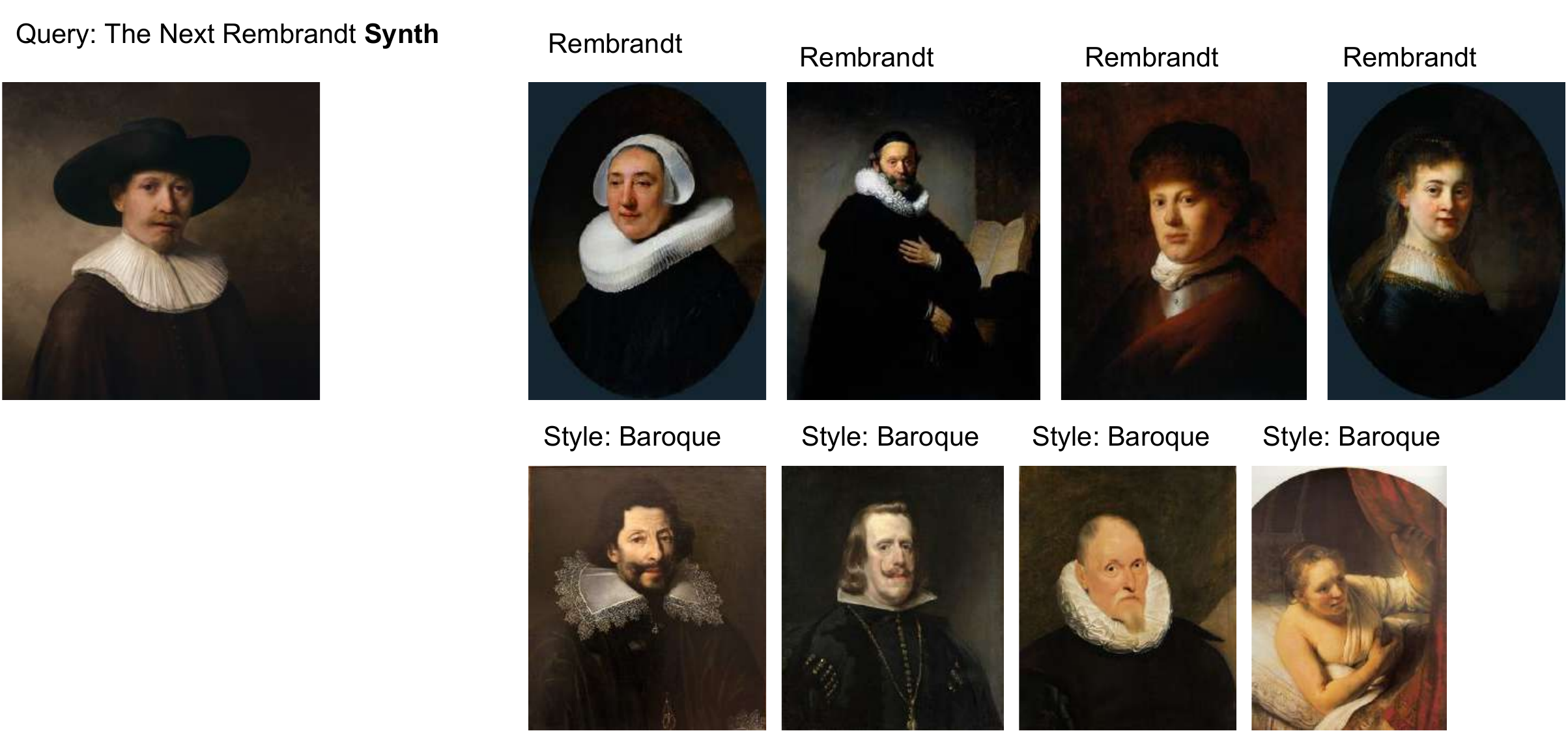}
\caption{Similarity results for a painting by The Next Rembrandt, belonging to the Baroque style. Similarity results: top four painting retrieved using the artist features (first row) and the style features (second row).}
\label{fig:example4}
\end{figure}

A final example is reported in Figure \ref{fig:example4}, where a painting from The Next Rembrandt (https://www.nextrembrandt.com/) is given as input to our system. The painting has been synthetically generated from data derived from 346 known paintings by Rembrandt, and was created from a deep, 18-month analysis of his work. A facial recognition algorithm learned Rembrandt’s techniques, pixel data helped the computer mimic brush strokes, and an advanced 3D printer brought the painting to life using 13 layers of ink. The portrait consists of 148 million pixels and is based on 168,263 fragments from Rembrandt’s portfolio. Interestingly, all the four most similar paintings retrieved by the system using the artist features are from Rembrandt, and all the the first four paintings retrieved using the style features belong to the Baroque style, that is the same style of the input painting.

\section{Conclusions}

In this work we tackled the problem of artist, style, and genre categorization of paintings. We proposed a new deep multibranch neural network to solve simultaneously all the three the problems in a multitask formulation.
The branches of the proposed network are fed with crops at different resolutions in order to gather clues from low-level texture details and exploit at the same time the coarse layout of the painting.
We proposed and compared two different cropping strategies: a random one, and one based on Spatial Transformer Networks. Furthermore, we have also experimented the injection in the proposed network of different hand-crafted features directly computed on the input images. The evaluation has been carried out on a new dataset originally sourced from wikiart.org and hosted by Kaggle, that we made suitable for artist, style and genre multitask learning. This dataset, that we named \DB{}, is composed of 100K paintings divided into 1508 artists, 125 styles and 41 genres.
We used \DB{} to evaluate and compare the effectiveness of the different variants of the proposed approach. The best solution, which exploits the STN cropping strategy and the injection of HOG features,
achieved accuracy levels of 56.5\%, 57.2\%, and 63.6\% on the tasks of artist, style and genre prediction respectively.
In order to facilitate a fair comparison with other methods in the state of the art, the \DB{} dataset along with the training and test splits used are made available as well as a web demo that makes it possible to interactively experience the proposed method (http://www.ivl.disco.unimib.it/activities/paintings/).

\section*{Acknowledgment}
The authors gratefully acknowledge the support of NVIDIA Corporation with the donation of the Titan X Pascal GPU used for this research.

\section*{References}

\bibliography{paintings_biblio}

\begin{thebibliography}{10}
\expandafter\ifx\csname url\endcsname\relax
  \def\url#1{\texttt{#1}}\fi
\expandafter\ifx\csname urlprefix\endcsname\relax\def\urlprefix{URL }\fi
\expandafter\ifx\csname href\endcsname\relax
  \def\href#1#2{#2} \def\path#1{#1}\fi

\bibitem{carneiro2012artistic}
G.~Carneiro, N.~P. da~Silva, A.~Del~Bue, J.~P. Costeira, Artistic image
  classification: An analysis on the printart database, in: European Conference
  on Computer Vision, Springer, 2012, pp. 143--157.

\bibitem{mensink2014rijksmuseum}
T.~Mensink, J.~Van~Gemert, The rijksmuseum challenge: Museum-centered visual
  recognition, in: Proceedings of International Conference on Multimedia
  Retrieval, ACM, 2014, p. 451.

\bibitem{khan2014painting}
F.~S. Khan, S.~Beigpour, J.~Van~de Weijer, M.~Felsberg, Painting-91: a large
  scale database for computational painting categorization, Machine vision and
  applications 25~(6) (2014) 1385--1397.

\bibitem{mao2017deepart}
H.~Mao, M.~Cheung, J.~She, Deepart: Learning joint representations of visual
  arts, in: Proceedings of the 2017 ACM on Multimedia Conference, ACM, 2017,
  pp. 1183--1191.

\bibitem{anwer2016combining}
R.~M. Anwer, F.~S. Khan, J.~van~de Weijer, J.~Laaksonen, Combining holistic and
  part-based deep representations for computational painting categorization,
  in: Proceedings of the 2016 ACM on International Conference on Multimedia
  Retrieval, ACM, 2016, pp. 339--342.

\bibitem{fichner2011foundations}
L.~Fichner-Rathus, Foundations of Art and Design: An Enhanced Media Edition,
  Cengage Learning, 2011.

\bibitem{widjaja2003identifying}
I.~Widjaja, W.~K. Leow, F.-C. Wu, Identifying painters from color profiles of
  skin patches in painting images, in: Image Processing, 2003. ICIP 2003.
  Proceedings. 2003 International Conference on, Vol.~1, IEEE, 2003, pp.
  I--845.

\bibitem{wikipaintings}
S.~Karayev, M.~Trentacoste, H.~Han, A.~Agarwala, T.~Darrell, A.~Hertzmann,
  H.~Winnemoeller, Recognizing image style, in: BMVC 2014 - Proceedings of the
  British Machine Vision Conference 2014, 2014, pp. 1--20.

\bibitem{peng2015cross}
K.-C. Peng, T.~Chen, Cross-layer features in convolutional neural networks for
  generic classification tasks, in: Image Processing (ICIP), 2015 IEEE
  International Conference on, IEEE, 2015, pp. 3057--3061.

\bibitem{chu2016deep}
W.-T. Chu, Y.-L. Wu, Deep correlation features for image style classification,
  in: Proceedings of the 2016 ACM on Multimedia Conference, ACM, 2016, pp.
  402--406.

\bibitem{puthenputhussery2016color}
A.~Puthenputhussery, Q.~Liu, C.~Liu, Color multi-fusion fisher vector feature
  for fine art painting categorization and influence analysis, in: Applications
  of Computer Vision (WACV), 2016 IEEE Winter Conference on, IEEE, 2016, pp.
  1--9.

\bibitem{puthenputhussery2016sparse}
A.~Puthenputhussery, Q.~Liu, C.~Liu, Sparse representation based complete
  kernel marginal fisher analysis framework for computational art painting
  categorization, in: European Conference on Computer Vision, Springer, 2016,
  pp. 612--627.

\bibitem{peng2016toward}
K.-C. Peng, T.~Chen, Toward correlating and solving abstract tasks using
  convolutional neural networks, in: Applications of Computer Vision (WACV),
  2016 IEEE Winter Conference on, IEEE, 2016, pp. 1--9.

\bibitem{banerji2016painting}
S.~Banerji, A.~Sinha, Painting classification using a pre-trained convolutional
  neural network, in: International Conference on Computer Vision, Graphics,
  and Image processing, Springer, 2016, pp. 168--179.

\bibitem{tan2016ceci}
W.~R. Tan, C.~S. Chan, H.~E. Aguirre, K.~Tanaka, Ceci n'est pas une pipe: A
  deep convolutional network for fine-art paintings classification, in: Image
  Processing (ICIP), 2016 IEEE International Conference on, IEEE, 2016, pp.
  3703--3707.

\bibitem{salehlarge}
B.~Saleh, A.~Elgammal, Large-scale classification of fine-art paintings:
  Learning the right metric on the right feature, International Journal for
  Digital Art History 2~(Oct) (2016) 71--93.

\bibitem{bianco2017deeppainting}
S.~Bianco, D.~Mazzini, R.~Schettini, Deep multibranch neural network for
  painting categorization, in: International Conference on Image Analysis and
  Processing, Springer, 2017, pp. 414--423.

\bibitem{huang2017fine}
X.~Huang, S.-h. Zhong, Z.~Xiao, Fine-art painting classification via
  two-channel deep residual network, in: Pacific Rim Conference on Multimedia,
  Springer, 2017, pp. 79--88.

\bibitem{chu2018image}
W.-T. Chu, Y.-L. Wu, Image style classification based on learnt deep
  correlation features, IEEE Transactions on Multimedia.

\bibitem{sharif2014cnn}
A.~Sharif~Razavian, H.~Azizpour, J.~Sullivan, S.~Carlsson, Cnn features
  off-the-shelf: an astounding baseline for recognition, in: Proceedings of the
  IEEE Conference on Computer Vision and Pattern Recognition Workshops, 2014,
  pp. 806--813.

\bibitem{bianco2015local}
S.~Bianco, D.~Mazzini, D.~Pau, R.~Schettini, Local detectors and compact
  descriptors for visual search: a quantitative comparison, Digital Signal
  Processing 44 (2015) 1--13.

\bibitem{peng2015framework}
K.-C. Peng, T.~Chen, A framework of extracting multi-scale features using
  multiple convolutional neural networks, in: 2015 IEEE International
  Conference on Multimedia and Expo (ICME), IEEE, 2015, pp. 1--6.

\bibitem{hentschel2016fine}
C.~Hentschel, T.~P. Wiradarma, H.~Sack, Fine tuning cnns with scarce training
  data—adapting imagenet to art epoch classification, in: Image Processing
  (ICIP), 2016 IEEE International Conference on, IEEE, 2016, pp. 3693--3697.

\bibitem{krizhevsky2012imagenet}
A.~Krizhevsky, I.~Sutskever, G.~E. Hinton, Imagenet classification with deep
  convolutional neural networks, in: Advances in neural information processing
  systems, 2012, pp. 1097--1105.

\bibitem{jaderberg2015spatial}
M.~Jaderberg, K.~Simonyan, A.~Zisserman, et~al., Spatial transformer networks,
  in: Advances in Neural Information Processing Systems, 2015, pp. 2017--2025.

\bibitem{he2016deep}
K.~He, X.~Zhang, S.~Ren, J.~Sun, Deep residual learning for image recognition,
  in: Proceedings of the IEEE Conference on Computer Vision and Pattern
  Recognition, 2016, pp. 770--778.

\bibitem{ioffe2015batch}
S.~Ioffe, C.~Szegedy, Batch normalization: Accelerating deep network training
  by reducing internal covariate shift, in: Proceedings of The 32nd
  International Conference on Machine Learning, 2015, pp. 448--456.

\bibitem{nair2010rectified}
V.~Nair, G.~E. Hinton, Rectified linear units improve restricted boltzmann
  machines, in: Proceedings of the 27th international conference on machine
  learning (ICML-10), 2010, pp. 807--814.

\bibitem{bianco2017large}
S.~Bianco, Large age-gap face verification by feature injection in deep
  networks, Pattern Recognition Letters 90 (2017) 36--42.

\bibitem{napoletano2018visual}
P.~Napoletano, Visual descriptors for content-based retrieval of remote-sensing
  images, International Journal of Remote Sensing 39~(5) (2018) 1343--1376.

\bibitem{mirmehdi2008handbook}
M.~Mirmehdi, Handbook of texture analysis, Imperial College Press, 2008.

\bibitem{lowe2004distinctive}
D.~G. Lowe, Distinctive image features from scale-invariant keypoints,
  International journal of computer vision 60~(2) (2004) 91--110.

\bibitem{novak1992anatomy}
C.~L. Novak, S.~Shafer, et~al., Anatomy of a color histogram, in: Computer
  Vision and Pattern Recognition, 1992. Proceedings CVPR'92., 1992 IEEE
  Computer Society Conference on, IEEE, 1992, pp. 599--605.

\bibitem{Pietikainen1996}
M.~Pietikainen, S.~Nieminen, E.~Marszalec, T.~Ojala, Accurate color
  discrimination with classification based on feature distributions, in:
  Proceedings of the 13th International Conference on Pattern Recognition,
  Vol.~3, 1996, pp. 833--838.

\bibitem{paschos2003image}
G.~Paschos, I.~Radev, N.~Prabakar, Image content-based retrieval using
  chromaticity moments, IEEE Transactions on Knowledge and Data Engineering
  15~(5) (2003) 1069--1072.

\bibitem{Bianconi2011}
F.~Bianconi, R.~Harvey, P.~Southam, A.~Fern\'andez, Theoretical and
  experimental comparison of different approaches for color texture
  classification, Journal of Electronic Imaging 20~(4).

\bibitem{Barilla2008}
M.~Barilla, M.~Spann, Colour-based texture image classification using the
  complex wavelet transform, in: Electrical Engineering, Computing Science and
  Automatic Control, 2008. CCE 2008. 5th International Conference on, 2008, pp.
  358 --363.

\bibitem{oliva2001modeling}
A.~Oliva, A.~Torralba, Modeling the shape of the scene: A holistic
  representation of the spatial envelope, Int'l J. Computer Vision 42~(3)
  (2001) 145--175.

\bibitem{Bianconi2007}
F.~Bianconi, A.~Fern\'andez, Evaluation of the effects of gabor filter
  parameters on texture classification, Pattern Recognition 40~(12) (2007) 3325
  -- 3335.

\bibitem{maenpaa2004classification}
T.~M\"{a}enp\"{a}\"{a}, M.~Pietik\"{a}inen, Classification with color and
  texture: jointly or separately?, Pattern Recognition 37~(8) (2004)
  1629--1640.

\bibitem{cusano2016combining}
C.~Cusano, P.~Napoletano, R.~Schettini, Combining multiple features for color
  texture classification, Journal of Electronic Imaging 25~(6) (2016)
  061410--061410.

\bibitem{cusano2014combining}
C.~Cusano, P.~Napoletano, R.~Schettini, Combining local binary patterns and
  local color contrast for texture classification under varying illumination,
  JOSA A 31~(7) (2014) 1453--1461.

\bibitem{cusano2013illuminant}
C.~Cusano, P.~Napoletano, R.~Schettini, Illuminant invariant descriptors for
  color texture classification, in: Computational Color Imaging, Vol. 7786 of
  Lecture Notes in Computer Science, 2013, pp. 239--249.

\bibitem{bianco2013robustness}
S.~Bianco, C.~Cusano, P.~Napoletano, R.~Schettini, On the robustness of color
  texture descriptors across illuminants, in: International Conference on Image
  Analysis and Processing, Springer, 2013, pp. 652--662.

\bibitem{chatzichristofis2008cedd}
S.~A. Chatzichristofis, Y.~S. Boutalis, Cedd: color and edge directivity
  descriptor: a compact descriptor for image indexing and retrieval, in:
  International Conference on Computer Vision Systems, Springer, 2008, pp.
  312--322.

\bibitem{junior2009trainable}
O.~L. Junior, D.~Delgado, V.~Gon{\c{c}}alves, U.~Nunes, Trainable
  classifier-fusion schemes: an application to pedestrian detection, in:
  Intelligent Transportation Systems, 2009.

\bibitem{yang2010bag}
Y.~Yang, S.~Newsam, Bag-of-visual-words and spatial extensions for land-use
  classification, in: Proc. of the Int'l Conf. on Advances in Geographic
  Information Systems, 2010, pp. 270--279.

\end{thebibliography}

\end{document}